\documentclass{article}

\usepackage{microtype}
\usepackage{graphicx}
\usepackage{subfigure}
\usepackage{booktabs} 
\usepackage{xcolor}

\usepackage{hyperref}
\usepackage{xurl}
\usepackage{amsmath,amsthm,amssymb}
\usepackage{cleveref}
\usepackage{enumitem}
\usepackage{array}

\usepackage[most]{tcolorbox}

\newtcolorbox{takeaway}{
    colback=gray!10,
    colframe=black,
    boxrule=1.5pt,
    arc=3pt,
    left=6pt,         
    right=6pt,        
    top=2pt,          
    bottom=2pt,       
    boxsep=2pt,       
    halign=justify,   
    fontupper=\small\itshape,
    width=\columnwidth
}

\usepackage[accepted]{mlsys2027}
\makeatletter
\renewcommand{\Notice@String}{Preprint.}
\makeatother

\mlsystitlerunning{Creating Power Flexibility in Datacenters via Query Routing}

\begin{document}
\twocolumn[
\mlsystitle{Creating Power Flexibility in Datacenters via Query Routing}

\mlsyssetsymbol{mentor}{*}
\mlsyssetsymbol{intern}{\textdagger}

\begin{mlsysauthorlist}
\mlsysauthor{Shivam Patel}{cmu,intern}
\mlsysauthor{Bingqing Chen}{bosch}
\mlsysauthor{Jared Fernandez}{cornell}
\mlsysauthor{Guannan Qu}{cmu,mentor}
\mlsysauthor{Gauri Joshi}{cmu,mentor}
\end{mlsysauthorlist}

\mlsysaffiliation{cmu}{Carnegie Mellon University}
\mlsysaffiliation{bosch}{Bosch Center for AI}
\mlsysaffiliation{cornell}{Cornell University}

\mlsyskeywords{power demand flexibility; query routing; efficient inference}

\vskip 0.3in

\begin{abstract}
The growing energy requirements of large language model inference highlight the need for datacenters to support safe grid operation by providing power demand flexibility, while maintaining service for interactive workloads.
This work investigates query routing as a mechanism to enable such capability by dynamically allocating queries across different language model instances without deferring requests. We develop a general analytical framework that connects query arrival rates and sequence length distributions to token iteration time, request residence time, serving capacity, and power consumption of model instances. Leveraging this framework, we characterize the feasible query routing allocations and the corresponding power demand flexibilities, enabling the selection of routing strategies to achieve desired power consumption behavior. We validate our framework across four GPU platforms on language models ranging from 1B to 70B parameters, obtaining 2.8\% mean absolute percentage error in estimating achievable power flexibility. We evaluate our approach on a real trace-driven simulation of a 10,000 H100 GPUs datacenter, exhibiting flexibility up to 16\% of total power consumption, with 90\% of the desired power reduction achieved within approximately 10 seconds. These results establish query routing as an effective mechanism for datacenter power demand flexibility and demonstrate the broader applicability of our proposed framework to datacenter level planning and orchestration.
\end{abstract}
]

\printAffiliationsAndNotice{%
  \textsuperscript{*}Equal mentoring.
  \textsuperscript{\textdagger}Work done during an internship at Bosch Center for AI.
}

\section{Introduction}
\label{sec:introduction}

Large language models (LLMs) have experienced rapid development and widespread adoption across question answering, code generation, conversational assistants, and large data applications. The growing demand and evolving use cases for these models require increasing compute capacity, driving the construction of larger GPU datacenters. Recent assessment estimates show that datacenters consumed approximately 485\,TWh globally in 2025, averaging a 55\,GW power demand, with AI-focused datacenter capacity increasing more than three-fold over the last 18 months \cite{iea2026keyquestions}. Furthermore, this demand is continuously growing as individual AI campuses are being developed at the gigawatt scale \cite{crusoe2025abilene_gigawatt,meta2026prometheus}.

The growth of datacenters is also changing the composition of overall electricity demand, with projections estimating that datacenters could account for 9.5--15.3\% of U.S. electricity consumption by 2030 \cite{smith2026datacenterupdate}. As these facilities contribute to an increasing proportion of grid demand, their operating characteristics must be incorporated into power-system planning and control to ensure safe and reliable grid operations. In particular, large and abrupt changes in datacenter demand can create imbalance between the generation and load, affecting the grid voltage and distribution frequency. For example, a transmission fault in July 2024 in Northern Virginia caused approximately 1.5\,GW of datacenter load to disconnect from the grid, resulting in sudden increase in voltages above normal operating levels, prompting corrective action \cite{nerc2025largeloadloss}. 

\begin{figure}
    \centering
    \includegraphics[width=\columnwidth]{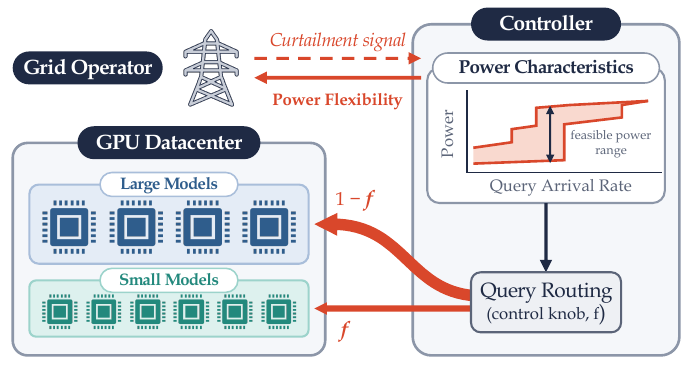}
    \vspace*{-6mm}
\caption{\textbf{Power demand flexibility through query routing.} Routing queries across large and small language models enables datacenters to adjust power demand for safe grid operation without dropping or deferring inference requests.}
\vspace*{-4mm}
    \label{fig:overview_diagram}
\end{figure}

\begingroup
\renewcommand{\thefootnote}{}
\footnotetext{Code available at: \href{https://github.com/patel-shivam/datacenter-power-flexibility-code-release}{GitHub}.}
\endgroup

Integrating datacenters into grid planning and control thus requires characterization of their load behavior, and one of the most important questions that grid operators care about is the demand flexibility of datacenters: \emph{how much power can datacenter reduce when requested, during periods of high grid loads, supply shortages, and other contingencies \cite{norris2025rethinking}?} 
Grid operators are progressively incorporating such power demand flexibility into agreements with datacenters, for example, Google reported contracting 1\,GW of datacenter demand response capacity with U.S. utilities in March 2026 \cite{google_demand_response_terrell2026milestone}. For datacenters serving inference workloads, adhering to these commitments requires reducing power consumption while continuing to serve inference request workloads.

Existing power reduction and curtailment strategies include deferring or rescheduling training and batch inference workloads \cite{carbon_aware_computing_datacenters,carbon_aware_scheduling_clusters_305953}, shifting computation across datacenters \cite{limitations_carbonaware_temporal_spatioal_scheduling_10.1145/3627703.3650079}, or adjusting GPU power consumption limits and clock frequencies \cite{zeus_gpu_energy_285082,perseus_energy_training_10.1145/3694715.3695970,stojkovic2025dynamollm,pals_hankendi2026palspowerawarellmserving}. Demand response frameworks further coordinate power curtailment across query workloads \cite{xing2023carbonrespondercoordinatingdemand}, and alternative on-site power sources and storage provide additional flexibility \cite{norris2025rethinking}. More discussion on prior work can be found in Appendix \ref{app:related_works}. Deferring query workloads is suitable for delay-tolerant tasks, whereas interactive real-time inference must satisfy latency service level objectives (SLOs). This motivates examining additional sources of power demand flexibility within the inference computational workload. 
In this work, we aim to achieve \emph{power demand flexibility for datacenters through query routing}.

Our key observation is that \emph{the choice of language model} for generating response to a query provides a controllable lever for modulating power consumption. 
Language query routing literature has demonstrated that smaller and efficient models match the response quality of larger models on a broad distribution of queries \cite{chen2024frugalgpt,ding2024hybrid,route_llm_ICLR2025_5503a7c6}. Such queries which are adequately served by varying language models present opportunities to dynamically control datacenter level power consumption. \Cref{fig:overview_diagram} describes our approach where we utilize these flexible query--model assignments and existing serving headroom at model instances to predict and control the power demand flexibility by routing queries without dropping or deferring them.

Our contributions are outlined as follows:
\begin{itemize}
    \item \textbf{Modeling LLM inference and power consumption.}
    We develop an analytical model (\Cref{sec:inference_power_model}) that estimates average token iteration time, mean request residence time, serving capacity and power consumption based on query arrival rates and query and response length distributions. This analytical model is validated across different hardware platforms, and language models on real and synthetic workloads. On H100 GPUs, our predictions across configurations demonstrate consistent accuracy with approximately $6\%$ error.     

    \item \textbf{Power flexibility through query routing.}
    We calculate the feasible routing decisions under query workloads and model instance capacity constraints, and quantify the resulting power demand flexibility. 
    This analysis establishes the effect of query workload and available serving headroom on power flexibility, and enables the selection of routing allocations to attain desired power consumption values. Evaluations substantiate our findings, with the predicted power demand flexibility within approximately 2.8\% mean error to the observed values (\Cref{sec:routing_power_control}).

    \item \textbf{Power demand flexibility at datacenter scale.}
    We simulate a deployment of 10{,}000 H100 GPUs using a week-long trace of inference queries on Azure platform \cite{stojkovic2025dynamollm}. Our approach demonstrates a mean power demand flexibility of up to 1.19\,MW (16\%) for queries which are indifferent to model choice. We also evaluate a sustained curtailment event over four hours, with our transient response achieving 90\% of the target power change in approximately 10\,s. Furthermore, we study the impact of GPU resource unavailability and downtime on the available demand flexibility (\Cref{sec:datacenter_power_control}). 
\end{itemize}

We organize remainder of the paper as follows. \Cref{sec:inference_power_model} introduces the analytical model for inference state and power consumption,  \Cref{sec:routing_power_control} derives the routing feasibility constraints and power control mechanism. \Cref{sec:datacenter_power_control} evaluates datacenter-scale power control and discusses broader applications, and \Cref{sec:conclusion} concludes our work. Related work and additional results are presented in \Cref{app:related_works,app:additional_results}.

\section{Modeling LLM Inference and Power Consumption}
\label{sec:inference_power_model}

To characterize the inference performance and power consumption during generation, we develop a general analytical model that translates the query workload into serving attributes. First, we model the relationship between the token generation time and the number of requests in each token iteration in \Cref{subsec:llm_inference_model}, enabling us to estimate request residence time and request throughput. Next, we establish the power consumption profile of a single model instance and use it to estimate the aggregate power consumption across multiple model instances based on query workloads in \Cref{subsec:power_consumption}. This analytical model of inference state and power consumption generalizes across domains and applications, and serves as the foundation for our subsequent analysis of demand flexibility through the job scheduling perspective. In addition to in-text description, \Cref{tab:power-notation} summarizes our notation used in the subsequent discussion.

\begin{figure*}
  \centering
  \includegraphics[width=\textwidth]{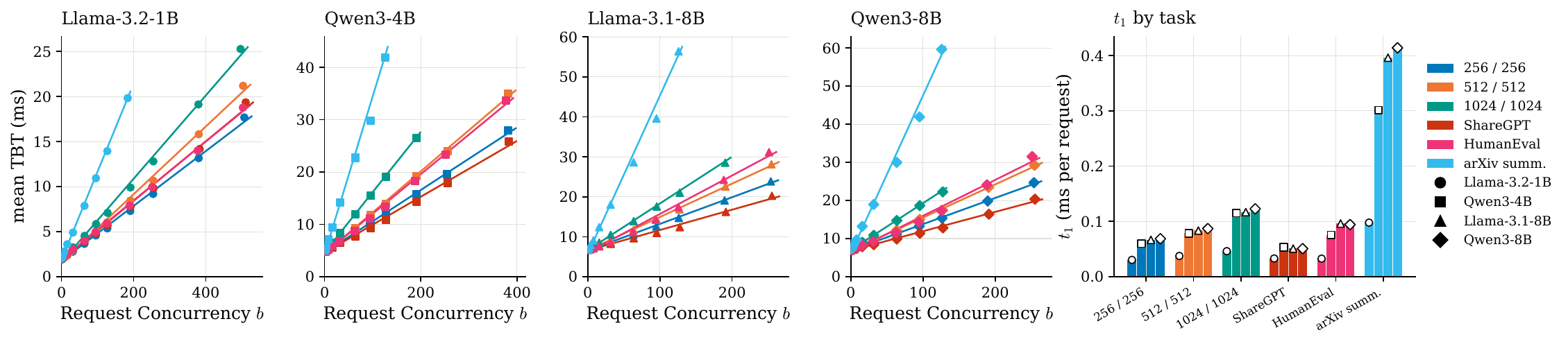}
  \vspace*{-2em}
  \caption{Measured and predicted mean token iteration times (TBT) as a function of number of concurrent requests across models and query length distributions. The linear estimator in \cref{eq:tbt_estimator} accurately captures the observed values ($R^2>0.99$ across all configurations). Larger models exhibit higher TBTs, and workloads with longer requests incur additional attention computation overhead.}
  \label{fig:tbt_plot}
  \vspace*{-1em}
\end{figure*}

\subsection{Modeling LLM Inference Serving}
\label{subsec:llm_inference_model}

\begin{takeaway}
Query arrival rates, query length distributions, and model instance profiling allows estimation of token iteration time, request residence time, throughput and request concurrency. 
\end{takeaway}

\paragraph{Workload and serving state.}
We consider a model instance with query arrival rate of $\lambda$ reqs/sec, with $\bar p,\bar d$ denoting the average prompt (prefill) and output (decode) lengths for queries respectively. Let $b$ represent the request concurrency, which includes queries/requests actively undergoing prefill or decode computations, distinguishing it from the number of tokens processed in an iteration. In practice, the request concurrency limit $b_{\max}$ is often controlled as a configurable limit based on memory or generation latency constraints.  Let $\tau(b)$ denote the average token iteration processing time, and $\widehat T(b)$ denote the estimated mean request residence time from beginning prefill computation to generation of last decode token.

\paragraph{Token Iteration Time.}
We estimate the average token processing time, also referred as time-between-tokens (TBT), for a given number of resident requests through the linear estimator 
\begin{equation}
  \tau(b)=t_0+t_1b,
  \qquad t_0>0,\quad t_1\geq0,
  \label{eq:tbt_estimator}
\end{equation}
where the intercept term $t_0$ captures the fixed computational overhead per token iteration, and the slope $t_1$ characterizes the additional computational load for each sequence for a given language model instance. 

\Cref{fig:tbt_plot} compares the measured and estimated token iteration processing times for different language models, each hosted on a single H100 GPU. Across all evaluated workloads and language model configurations, our proposed estimator in \cref{eq:tbt_estimator} achieves an $R^2$ value of over $0.99$, demonstrating a strong fit to the measured values. We observe that larger models with more parameters lead to longer batch processing times, and query distributions with longer requests incur higher TBT due to increasing per-token attention computation costs. 
The accurate TBT estimator provides a reliable basis for estimating mean request residence time and request throughput, which we subsequently use to quantify the aggregate power consumption across query workloads.

\begin{figure}[h]
  \centering
  \includegraphics[width=\columnwidth]{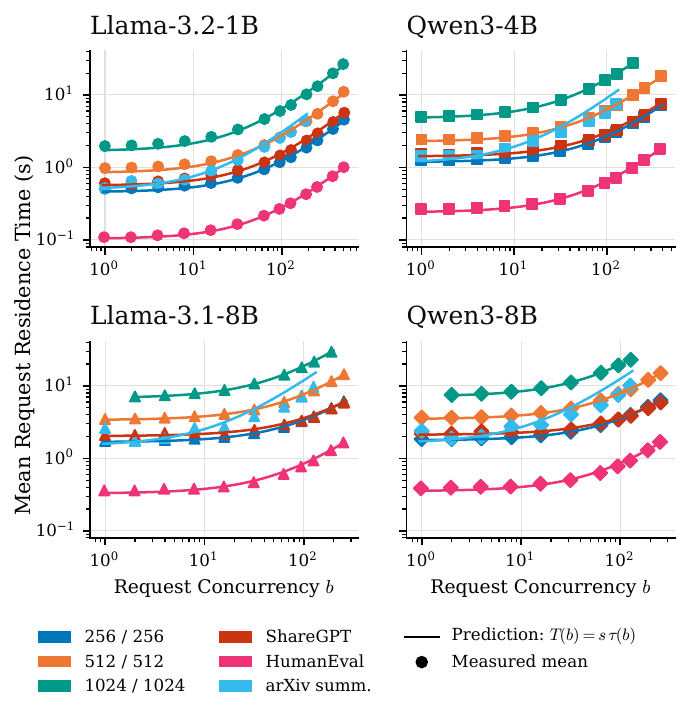}
  \vspace*{-2em}
  \caption{Predicted versus measured mean request residence time across configurations. The analytical estimator (\cref{eq:e2e_generation_time}) captures the increase in request duration with concurrency, achieving an overall $5.5\%$ MAPE, demonstrating reliable estimation of mean request residence time.}
  \label{fig:e2e_generation_time}
  \vspace*{-1em}
\end{figure}

\paragraph{Request Residence Time.}
We estimate the mean request residence time, including prefill and decode tokens, by approximating the number of token iterations required for end-to-end generation as
\begin{equation}
s=\bar d+\frac{\bar p}{\text{chunk}_{\mathrm{pre}}},
\qquad
\widehat T(b)=s\tau(b).
\label{eq:e2e_generation_time}
\end{equation}
Here $\bar d$ captures the average number of sequential decode iterations, and $\bar p/\text{chunk}_{\mathrm{pre}}$ approximates the number of token batches for prefill computation, where $\text{chunk}_{\mathrm{pre}}$ represents the number of prompt tokens processed in parallel in each prefill iteration. Thus, $s$ approximates the total number of token iterations required for generation, and $\widehat T(b)$ estimates the mean request residence time with $b$ active requests. \Cref{fig:e2e_generation_time} depicts the mean request residence times based on the request concurrency values ($b$), demonstrating a $5.5\%$ mean absolute percentage error (MAPE) over the examined configurations.

\begin{figure}[h]
  \centering
  \includegraphics[width=\columnwidth]{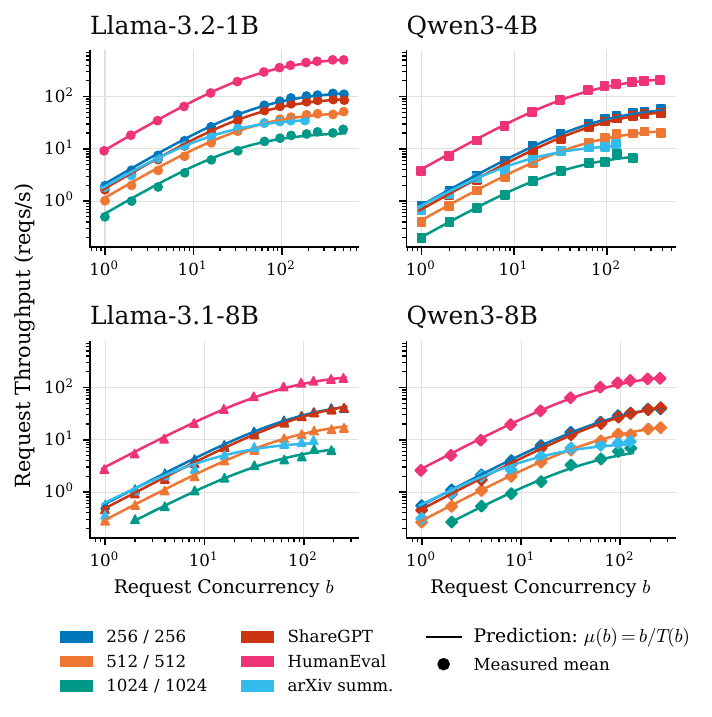}
  \vspace*{-2em}
  \caption{Predicted and measured request throughput as a function of request concurrency across configurations. Throughput increases with concurrency but exhibits diminishing returns as the token iteration times grow. The analytical estimator (\cref{eq:throughput_estimator}) achieves an overall MAPE of $5.8\%$, accurately capturing the concurrency-dependent request throughput of model instances.}
  \label{fig:generation_throughput_vs_concurrency}
  \vspace*{-1em}
\end{figure}

\paragraph{Request Throughput and Concurrency.}
We use the Little’s law to estimate the mean request concurrency as request throughput times the mean request residence time. Thus, the request throughput at concurrency $b$ is
\begin{equation}
\widehat\mu(b)
=\frac{b}{\widehat T(b)}
=\frac{b}{s(t_0+t_1b)},
\quad
\widehat C(b_{\max})=\widehat\mu(b_{\max}),
\label{eq:throughput_estimator}
\end{equation}
where $\widehat C(b_{\max})$ denotes the estimated serving capacity at the maximum concurrency limit, and $\widehat\mu(0)=0$. Increasing the concurrency $b$ raises request throughput with diminishing returns, along with increasing the mean token iteration processing times. \Cref{fig:generation_throughput_vs_concurrency} illustrates the predicted request throughput, with an overall $5.8\%$ MAPE to the observed measurements.

\begin{figure}[h]
  \centering
  \includegraphics[width=\columnwidth]{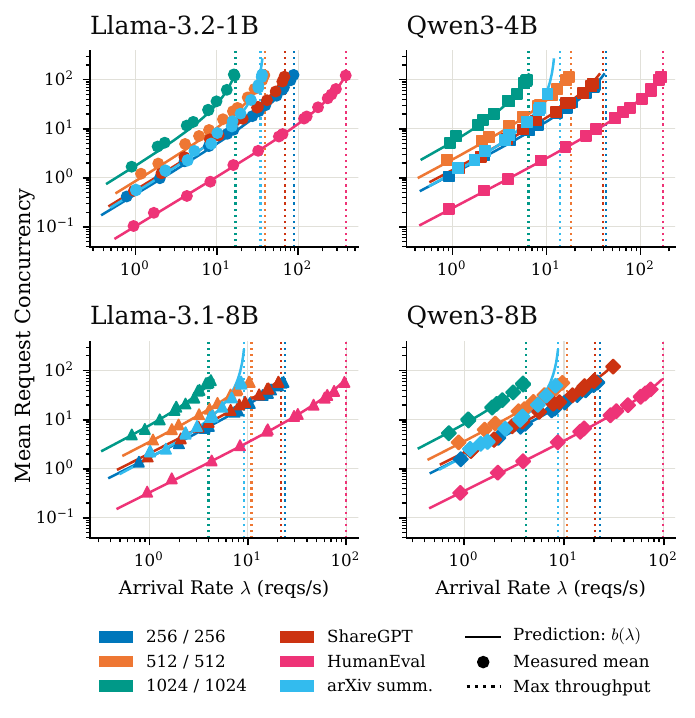}
  \vspace*{-2em}
\caption{Estimating request concurrency (\cref{eq:occupancy_estimator}) from query arrival rate, demonstrating an overall MAPE of $9.2\%$ over feasible operating points. The predicted request concurrency count connects the query arrival rate to the model instance's inference and power consumption.}
  \label{fig:occupancy_vs_arrival_rate}
  \vspace*{-1em}
\end{figure}

Under stable operation without growing queues, the request throughput $\widehat\mu(b)$ matches the arrival rate $\lambda$ reqs/s. Rewriting the request throughput (\cref{eq:throughput_estimator}) gives:
\begin{equation}
\widehat b(\lambda)
=\frac{\lambda s t_0}{1-\lambda s t_1},
\qquad 0\leq\lambda\leq\widehat C(b_{\max}).
\label{eq:occupancy_estimator}
\end{equation}
This expression connects the query arrival rate in req/s to the operating state at the model instance, permitting the inference state calculations for token iteration time (\cref{eq:tbt_estimator}) and mean request residence time (\cref{eq:e2e_generation_time}). The predicted mean request concurrency depicts an average $9.2\%$ MAPE from the observed values, as presented in \Cref{fig:occupancy_vs_arrival_rate}. 

\subsection{Modeling Power Consumption}
\label{subsec:power_consumption}

\begin{takeaway}
Estimating power based on request concurrency connects query arrival rate to model instance power consumption, allowing joint analysis over multiple models. 
\end{takeaway}

\paragraph{Model Instance power profiles.}
We model the mean board power consumed by a model instance as a function of request concurrency by integrating the serving state developed above with power consumption.
For a given model instance and query workload, we represent the power as
\begin{equation}
  P(b)=P_{\mathrm{idle}}+P_{\mathrm{dyn}}(b),
  \qquad P_{\mathrm{dyn}}(0)=0,
  \label{eq:model_instance_power}
\end{equation}
where $P_{\mathrm{idle}}$ denotes the measured idle power consumption without any active requests (approximately $10-20\%$ of rated maximum power consumption). The dynamic component $P_{\mathrm{dyn}}(b)$ captures the additional power consumed during active inference and its dependence on the concurrency $b$. For each paired model and hardware configuration, we calibrate the dynamic component based on observed power consumption values. 

\begin{figure}[h]
  \centering
  \includegraphics[width=\columnwidth]{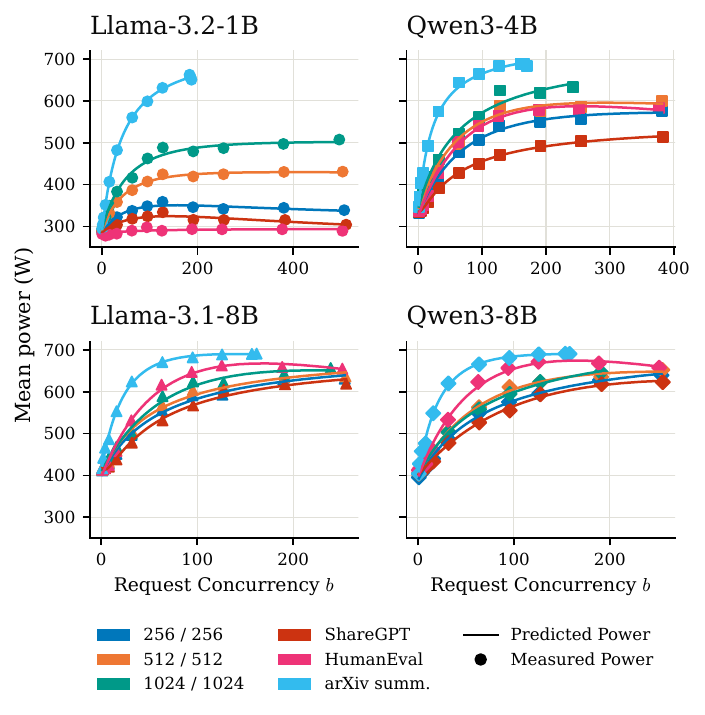}
  \vspace*{-2em}
\caption{Measured and predicted mean GPU board power as a function of request concurrency across configurations. Power consumption increases sharply when transitioning from idle to active inference, followed by a more gradual increase as concurrency grows. The diminishing incremental power at higher concurrency suggests opportunities for reducing aggregate power consumption through request consolidation across model instances.}
  \label{fig:power_consumed_vs_occupancy}
  \vspace*{-1em}
\end{figure}

\Cref{fig:power_consumed_vs_occupancy} illustrates the measured and estimated power consumption values from \cref{eq:model_instance_power}. The total consumed power typically increases sharply from the idle power, followed by a more gradual rise with increasing number of requests served in parallel ($b$). Across all examined configurations in \Cref{fig:power_consumed_vs_occupancy}, the analytical power model in \cref{eq:model_instance_power} achieves a median held-out RMSE of 5.2\,W.

\paragraph{Power Consumption and Query Arrival Rate.}
Utilizing the occupancy estimate in \cref{eq:occupancy_estimator} with the model instance power profile (\cref{eq:model_instance_power}) results in 
\begin{equation}
  \widehat P(\lambda)
    =P\bigl(\widehat b(\lambda)\bigr),
  \qquad
  0\leq\lambda\leq\widehat C(b_{\max}).
  \label{eq:model_instance_power_from_arrivalrate}
\end{equation}
This permits the estimation of power consumption directly from the arrival rate of queries, eliminating the need to measure average number of requests served in parallel at deployment. 

\paragraph{Generalization to Multiple Instances.}
We extend the examination of a single model instance to multiple instances by aggregating power predictions across instances. Consider a pool of model instances $j$ with $n_j$ homogeneous instances serving a total query arrival rate of $\lambda_j$ req/s. The load distribution and scheduling policies determine how requests are distributed among model instances, resulting in predicted request concurrency values for which the aggregate request throughput equals the query arrival rate:
\begin{equation}
\sum_{i=1}^{n_j}
\widehat\mu_j(\widehat b_{j,i})=\lambda_j,
\qquad
0\leq\widehat b_{j,i}\leq b_{\max,j}.
\label{eq:multiple_model_throughput}
\end{equation}
The total power consumption is obtained by aggregating the predicted power of all model instances:
\begin{equation}
\widehat P_{\mathrm{total}}
=\sum_j\sum_{i=1}^{n_j}P_j(\widehat b_{j,i}).
\label{eq:multiple_model_power}
\end{equation}
This formulation captures the varying choices of model instances and job scheduling policies, to predict the total power consumption without requiring additional calibration.

\paragraph{Request Scheduling across Identical Instances.}
Request scheduling and serving states across homogeneous model instances influence the aggregate power consumption. As observed in \Cref{fig:power_consumed_vs_occupancy}, the model instance power increases sharply when transitioning from idle to active execution, followed by a more gradual increase with increasing number of concurrently served requests. This implies that consolidating requests onto fewer model instances, up to their concurrency limit $b_{\max}$, results in lower total power consumption compared to distributing requests evenly across all instances. Thus, request consolidation avoids the substantial idle-to-active power increase incurred for  additional model instances, while benefiting from the diminishing incremental power consumption with higher request concurrency.

\paragraph{Experimental setup.}
Our presented evaluation of the language generation and power consumption analytical model in this section (\Cref{fig:tbt_plot,fig:e2e_generation_time,fig:generation_throughput_vs_concurrency,fig:occupancy_vs_arrival_rate,fig:power_consumed_vs_occupancy}) utilizes vLLM serving framework \cite{vllm_10.1145/3600006.3613165} on H100 GPUs. We examine models from Llama \cite{grattafiori2024llama3herdmodels} and Qwen \cite{qwen3_yang2025qwen3technicalreport} opensource families, with queries sampled from the ShareGPT \cite{sharegpt_dataset_vicuna2023}, HumanEval \cite{humaneval_dataset_chen2021evaluatinglargelanguagemodels}, arXiv Summarization \cite{arxiv_summarization_dataset_cohan-etal-2018-discourse} datasets along with fixed prompt/response length configurations of 256/256, 512/512, and 1024/1024 tokens. We show additional results on other hardware and models served in tensor parallel configuration in~\Cref{app:additional_results}.

\section{Power Control via Query Routing}
\label{sec:routing_power_control}

The analytical model developed in \Cref{sec:inference_power_model} connects the query workload to the inference state and power consumption by model instances. Building on this general analytical model, we examine how routing queries across model instances hosting different language models enables control over total power consumption. We first motivate this approach by discussing how query routing can exploit differences in model capabilities while preserving generative performance. We then formalize feasible routing allocations and capacity constraints, characterize the range of power control enabled by feasible routing decisions, and finally present a technique for selecting query-model instance assignments that meet specified power targets. 

\paragraph{Query Routing for Power Control.}
Practical applications of language models span a variety of use cases, hence the query difficulty and model capability requirements vary greatly. A large fraction of queries/requests do not require extensive capabilities of language models, and can instead be 
served equally well by smaller and more efficient language models. Prior work in language query routing shows that selectively routing queries to smaller models can preserve the generative performance of large language models \cite{chen2024frugalgpt,ding2024hybrid,route_llm_ICLR2025_5503a7c6}. This allows us to control total power consumption by scheduling inference queries across model pools, without affecting generative performance. 

We primarily study query routing across small and large model pools as an example of this principle, which broadly applies to dynamic inference configurations such as test-time scaling through reasoning token budgets \cite{snell2025scaling}, multi-response aggregation \cite{ding2025bestroute} etc.

\subsection{Formulation and Setup}
\label{subsec:routing-capacity}

\begin{takeaway}
What feasible routing allocations for queries do existing model instances and their serving capacities permit?
\end{takeaway}

Consider a pool of large and a pool of small model instances. For our analysis, we consider queries classified into three routing classes, where each class denotes the model assignment choice for a given query. Large-only and small-only queries demand generation from respective model pools due to reasoning capability or latency requirements, whereas flexible queries can be served by either pool. Let their workload share in fraction of total queries be $\sigma_L, \sigma_S,$ and $\sigma_F$ respectively, where $\sigma_L\!+\!\sigma_S\!+\!\sigma_F\!=\!1$. We assume the aforementioned query assignment is derived from one of the existing query routing methods and focus our scope on characterizing the resulting capacity constraints and power demand flexibility.

For total query arrival rate $\Lambda$ reqs/s, let $f\in[0,1]$ denote the
fraction of flexible requests assigned to the pool of small language model instances.
The total query arrival rate at large and small model instances is thus:
\begin{equation}
\begin{aligned}
  \lambda_L(f)
    &= \Lambda[\sigma_L+\sigma_F(1-f)],\\
  \lambda_S(f)
    &= \Lambda[\sigma_S+\sigma_F f].
\end{aligned}
\label{eq:small_large_arrival_rates}
\end{equation}
Varying the fraction $f$ controls the distribution of flexible query workload across small and large language model pools along with existing large- and small-only query workloads.

Let $n_L,n_S$ denote the number of large and small model instances, and $C_L,C_S$ denote their per-instance request throughput (reqs/s). To ensure stable workload at each language model pool, we constrain the routing fraction $f$ to satisfy serving capacity constraints as 
\begin{equation}
\begin{aligned}
  f_{\mathrm{lo}}&=\max\!\left(0,\frac{\Lambda(\sigma_L+\sigma_F)-n_LC_L}{\Lambda\sigma_F}\right),\\
  f_{\mathrm{hi}}&=\min\!\left(1,\frac{n_SC_S-\Lambda\sigma_S}{\Lambda\sigma_F}\right).
\end{aligned}
\label{eq:feasible_routing_fraction_range}
\end{equation}
for $\Lambda>0$ and $\sigma_F>0$. The lower bound represents the minimum fraction of flexible queries that must be assigned to the small language model pool to prevent overloading large language model pool, and vice-versa for the upper bound on routing fraction $f$. Accordingly, the feasible routing set is $\mathcal F=[f_{\mathrm{lo}},f_{\mathrm{hi}}]$ when
$f_{\mathrm{lo}}\leq f_{\mathrm{hi}}$; otherwise the provisioned model instances lack sufficient capacity to serve the existing query arrival rate under any configuration. 

As the total query arrival rate ($\Lambda$) increases, the extra headroom at existing model instances for serving flexible queries decreases, constraining the feasible routing decisions. We examine routing decisions on a four model instance pool with three Llama-3.1-8B and one Llama-3.2-1B instance, with $21.75$ and $69$ req/s serving capacity respectively for queries sampled from the ShareGPT dataset ($\sigma_F\!=\!1$). As we increase the total query arrival rate from $66$ to $119$ reqs/s, we observe that the feasible routing range shrinks from $[0.01,1]$ to $[0.45,0.58]$, indicating greater congestion and reducing headroom at existing model instances.

\begin{figure}[h]
  \centering
  \includegraphics[width=\columnwidth]{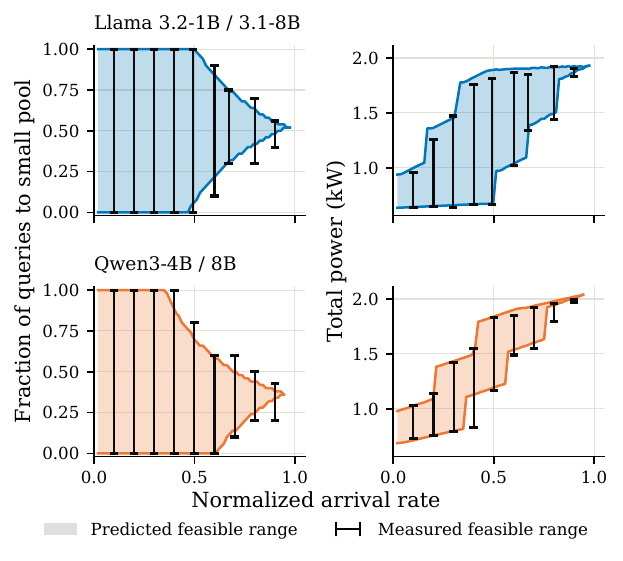}
  \vspace*{-2em}
  \caption{Capacity-feasible routing decisions (left, \cref{eq:feasible_routing_fraction_range}) and
  corresponding total-power consumption (right, \cref{eq:routing_total_power}) across query arrival rates and model--workload configurations.
  Shaded regions show analytical predictions, bars depict measured ranges.}
  \label{fig:feasible_routing_power}
\end{figure}

\begin{figure*}[t]
  \centering
  \includegraphics[width=\textwidth]{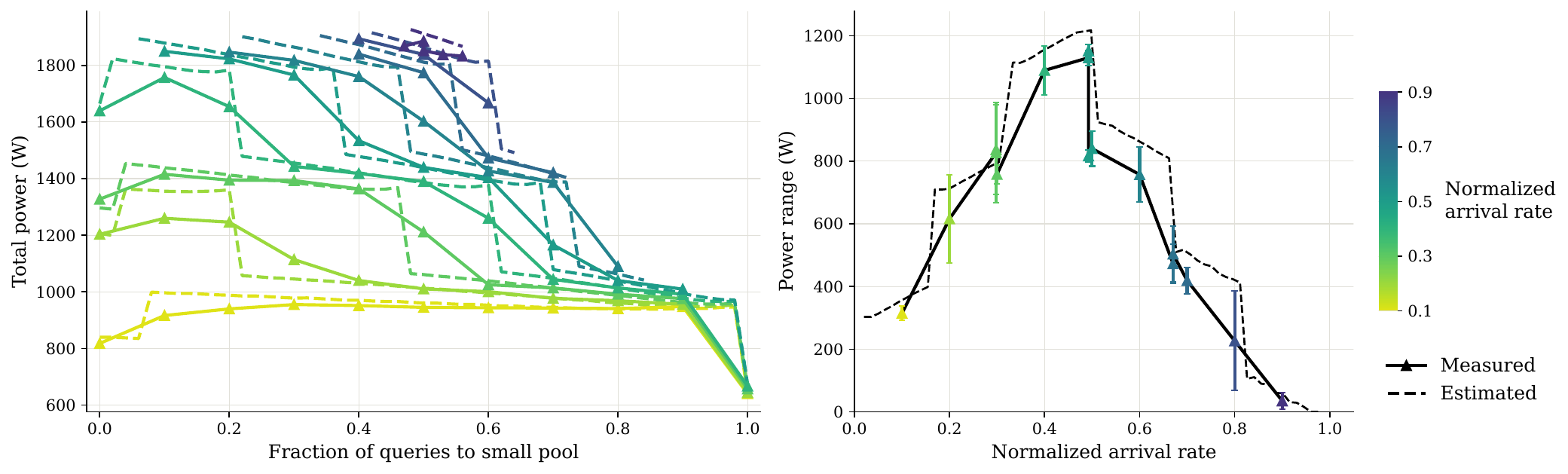}
  \vspace*{-2em}
  \caption{Total power consumption via query routing for the three Llama-8B and single Llama-1B model instances with queries sampled from ShareGPT.
  Left: measured and predicted total power (\cref{eq:routing_total_power}) versus the fraction of
  flexible requests ($f$) assigned to the small model instance, with varying query arrival rates. 
  Right: measured and predicted power flexibility range (\cref{eq:feasible_routing_power}) over feasible allocations.}
  \label{fig:power_range_vs_load}
  \vspace*{-0.5em}
\end{figure*}

The LLM inference model presented in (\cref{eq:tbt_estimator,eq:e2e_generation_time,eq:throughput_estimator}) provides us with model instance capacities which we utilize in \cref{eq:feasible_routing_fraction_range} to analytically determine the feasible routing decisions, and compare them with empirically obtained values. \Cref{fig:feasible_routing_power} demonstrates the accuracy of the estimated feasible routing fraction with respect to the empirically observed values, as the total query arrival rate at model instances increases. Next, we present the effect of feasible routing decisions on power demand flexibility.

\subsection{Power Flexibility through Query Routing}
\label{subsec:routing-envelope}

\begin{takeaway}
What is the available power demand flexibility for a given query arrival rate and hardware configuration?
\end{takeaway}

For a chosen pool of language model instances, let $\widehat P_j(\lambda_j)$ denote the predicted aggregate power consumption from \cref{eq:multiple_model_throughput,eq:multiple_model_power}. Substituting the designated query arrival rate at small and large model pools from \cref{eq:small_large_arrival_rates} results in 
\begin{equation}
  \widehat P_{\mathrm{total}}(\Lambda,f)
    =\widehat P_L\bigl(\lambda_L(f)\bigr)
     +\widehat P_S\bigl(\lambda_S(f)\bigr).
  \label{eq:routing_total_power}
\end{equation}
Query routing affects the power consumed through both the request concurrency as well as the number of active model instances. Provisioned model instances without active requests continue to contribute their idle power.
For a feasible range of routing decisions, we define the predicted power
consumption, $[\widehat P_{\min}(\Lambda),  \widehat P_{\max}(\Lambda)]$, and available flexibility, $\widehat B(\Lambda)$, as
\begin{equation}
  \begin{aligned}
    \widehat P_{\min}(\Lambda)
      &=\inf_{f\in\mathcal F}
        \widehat P_{\mathrm{total}}(\Lambda,f),\\
    \widehat P_{\max}(\Lambda)
      &=\sup_{f\in\mathcal F}
        \widehat P_{\mathrm{total}}(\Lambda,f),\\
    \widehat B(\Lambda)
      &=\widehat P_{\max}(\Lambda)
        -\widehat P_{\min}(\Lambda).
  \end{aligned}
  \label{eq:feasible_routing_power}
\end{equation}
The power flexibility quantifies the available power control at a given query arrival rate and deployment configuration. 
\Cref{fig:feasible_routing_power} illustrates the analytical predictions for power range with measured routing sweeps for four model instances, with three large model instances and one small model instance for Llama 3.2-1B/3.1-8B and Qwen3-4B/8B models. Across examined query workloads and model instance configurations, the feasible routing decision estimates are within $3.3\%$ mean absolute error to the empirically observed values, and the predicted power range closely tracks the measured values with approximately $2.8\%$ mean error.

\paragraph{Power Flexibility and Query Arrival Rate.}
At low query arrival rates, the presented computational workload limits the total power consumption across model instances. Alternatively, high query arrival rates saturate existing model instances, reducing the headroom for power flexibility by query routing. Therefore, intermediate arrival rates demonstrate the highest power demand flexibility for a pool of model instances. 

\Cref{fig:power_range_vs_load} depicts the total power consumed and the power demand flexibility across four model instances (three Llama3.1-8B, one Llama3.2-1B and $\sigma_F\!=\!1$) as the query arrival rate varies from $10\%$ to $90\%$ of the measured capacities. As the query arrival rate increases, the feasible set of routing decisions shrinks (also demonstrated in \Cref{fig:feasible_routing_power}), and the power consumed at a given routing fraction increases due to higher computational demand from the existing model instances. Additionally, the power demand flexibility is largest at intermediate query arrival rates, and decreases at higher or lower arrival rates.

\subsection{Selecting Routing Allocations for Power Targets}
\label{subsec:power-target-selection}

\begin{takeaway}
    How do we select the desired query routing allocation?
\end{takeaway}

Given a target aggregate power consumption $P_{\mathrm{target}}$, we leverage the analytical power model (\cref{eq:routing_total_power}) to identify a capacity-feasible routing allocation that minimizes the deviation from the desired power target:
\begin{equation}
    f^* = \arg\min_{f \in \mathcal{F}}
    \left| \widehat{P}_{\mathrm{total}}(\Lambda, f)
    - P_{\mathrm{target}} \right|.
    \label{eq:power_target_routing}
\end{equation}
We search for the required routing decision $f^*$ over the feasible routing interval $\mathcal{F}$, resulting in an allocation which provides the closest achievable power level under the existing workload and model instance configurations. As query arrival rates, feasible routing decisions or available serving capacities change, the allocation can be recomputed using updated estimates from the analytical model.

Routing decisions are applied to new arriving queries, while the existing responses already being generated at model instances are unaffected. Consequently, the total power consumption does not change instantaneously following a change in the routing decisions. As discussed in \Cref{app:cross-gpu-validation}, the time required for these transitions depends on the completion time of outstanding requests, which is roughly in the order of seconds \cite{vllm_10.1145/3600006.3613165}.

\paragraph{Extension to Multiple Models and Inference Budgets.}
Our formulation extends to multiple language model pools by replacing the scalar routing fraction $f$ with an allocation vector which specifies query assignments across eligible model instance pools. Feasible allocations must satisfy respective capacity constraints, and subsequently the aggregate power and flexibility across model instances is estimated through  \cref{eq:routing_total_power,eq:feasible_routing_power}. 
This principle can also be extended to inference configurations with varying test-time compute budgets, including but not limited to reasoning-token budgets and multiple sampled responses. Since the benefits of additional computation depend on query difficulty~\cite{snell2025scaling}, adaptive inference configurations can provide additional power flexibility while preserving generative quality~\cite{ding2025bestroute}.

In the next section, we discuss how query routing across model instances can support power demand flexibility and management at datacenter scale.
\section{Datacenter-Scale Power Control}
\label{sec:datacenter_power_control}

Increasingly large datacenter workloads and their contribution to grid power demand has resulted in exploration of datacenter power control for supporting grid operations \cite{google_demand_response_terrell2026milestone}. In this section, we study query routing across heterogeneous language models as a source of demand flexibility for maintaining grid compliance and versatility. We build upon the mechanisms developed in \Cref{sec:routing_power_control} and examine them at scale to demonstrate the practical import of our approach. In \Cref{subsec:datacenter_setup}, we describe the setup and design choices for our analysis, \Cref{subsec:datacenter_results} presents empirical results on a real-world workload trace, and \Cref{subsec:datacenter_broader_applications} comments on broader applications and scope.

\subsection{Empirical Setup}
\label{subsec:datacenter_setup}

We simulate a datacenter-scale deployment with $10,000$ Nvidia H100 GPUs, divided in a $3\!:\!1$ ratio between large and small model instances respectively. We examine Llama-3.1-70B \cite{grattafiori2024llama3herdmodels} served with tensor parallel configuration across four GPUs as the large model instance, and Mistral-24B \cite{mistralai2025magistral} served on a single GPU as the small model instance (\Cref{app:experimental-setup}). Our examined empirical results build on the inference and power consumption modeling for these language models as presented in \Cref{sec:inference_power_model}.

To evaluate our approach under practical scenarios, we utilize a week-long Azure LLM chat conversation trace \cite{stojkovic2025dynamollm}, with mean prompt and response lengths of $1630$ and $105$ tokens respectively. For this workload, the measured per-instance throughput is $7$ reqs/s for the larger model instance (Llama-70B), and $6.25$ reqs/s for the smaller model instance (Mistral-24B). We scale the query arrival rates such that the peak five-minute mean arrival rate over the week reaches $60\%$ of the total aggregate serving capacity across all model instances.

For the baseline operation, the query proportions for large and small model pools follow their aggregate serving capacities, i.e., $(\sigma_L + \sigma_F(1-f)) / (\sigma_S + \sigma_F f) = n_LC_L/n_SC_S$, following \cref{eq:small_large_arrival_rates}. Under this allocation, both the large and small model pools operate at the same fraction of their respective capacities. We vary the fraction of flexible queries ($\sigma_F$) in the total workload to evaluate the resulting power demand flexibility and available curtailment. For each $\sigma_F$ value, the remaining workload among large-only ($\sigma_L$) and small-only ($\sigma_S$) queries is maintained in proportion to their corresponding capacities, such that $\sigma_L/\sigma_S = n_LC_L/n_SC_S$. During power curtailment events, flexible queries are progressively directed toward the smaller language model's instances, subject to available headroom and feasibility constraints defined in \cref{eq:feasible_routing_fraction_range}.

In datacenter establishments, components such as thermal regulation, networking equipment, storage systems etc consume additional power along with GPUs \cite{power_management_opportunities_10.1145/3620666.3651329,stojkovic2024greenerllmsbringingenergyefficiency,tapas_10.1145/3676641.3716025}. Furthermore, power conversion and distribution losses introduce overheads which are captured through PUE (Power Usage Effectiveness) metrics. For our analysis, we model these workloads through an additional $300$W per GPU, and an aggregate PUE value of $1.2$, implying that the actual datacenter facility power is
\begin{equation}
  \widehat P_{\mathrm{facility}}(\Lambda,f)
    =1.2\left[
      \widehat P_{\mathrm{total}}(\Lambda,f)
      +3.0\,\mathrm{MW}
    \right].
  \label{eq:datacenter_facility_actual_power}
\end{equation}
We examine three aspects of datacenter power demand flexibility through the aforementioned setup. First, we quantify the trends in available power demand flexibility over the week-long real workload trace. Second, we simulate a four-hour curtailment event to assess the achievable reduction and the observed response times. Third, we analyze how changes in GPU availability, including allocation to training jobs or temporary outages, affect the available flexibility.

\begin{figure}[h]
  \centering
  \includegraphics[width=\columnwidth]
    {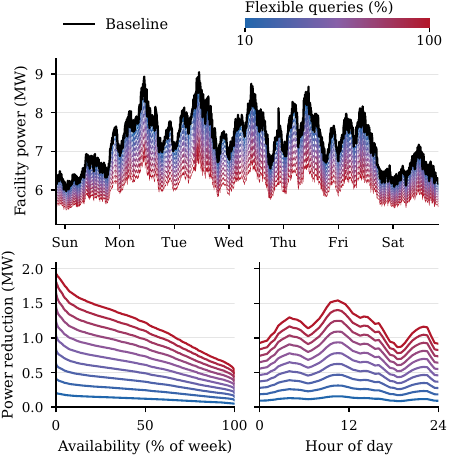}
  \vspace*{-2em}
\caption{Datacenter power demand flexibility over week-long Azure LLM workload trace.
Top: Baseline facility power and minimum achievable power under varying fractions of flexible queries.
Bottom: Distribution and hourly variation of available power curtailment, highlighting its dependence on query workload.}
  \label{fig:datacenter_week_plot}
  \vspace*{-0.5em}
\end{figure}

\subsection{Power Flexibility under Real Workloads}
\label{subsec:datacenter_results}

\paragraph{Available power flexibility over time.}
Query arrival rate varies with time, changing both the datacenter facility power consumption and the demand flexibility achievable through routing. We therefore examine a real-world week-long trace to characterize the distribution of available power control throughout the week and hourly averages during the day.

\Cref{fig:datacenter_week_plot} (top) depicts the datacenter facility power consumption (\cref{eq:datacenter_facility_actual_power}) throughout the week, along with available power curtailment distribution values with varying fraction of flexible query workloads. 
The baseline power consumption varies from $6$ to $9.1$\,MW, with a mean of $7.3$\,MW. With half of the query workload suited to both large and small model instances ($\sigma_F\!=\!0.5$), query routing across language models results in a power curtailment of $0.6$\,MW, which increases to $1.19$\,MW for $\sigma_F\!=\!1$, corresponding to $16\%$ of the mean facility power.

Furthermore, the availability varies across the week based on distribution of the query arrival rate (bottom \Cref{fig:datacenter_week_plot}). With half of all queries as flexible, atleast $0.32$\,MW power curtailment is available for $95\%$ of the week. Additionally, high query arrival rates during evening hours result in lower power flexibility of $0.49$\,MW as compared to the weekly mean of $0.6$\,MW. 
Thus, the power demand flexibility benefits from larger headroom during periods of intermediate query arrival rates, and improves with more flexible queries.

\begin{figure}[t]
  \centering
  \includegraphics[width=\columnwidth]
    {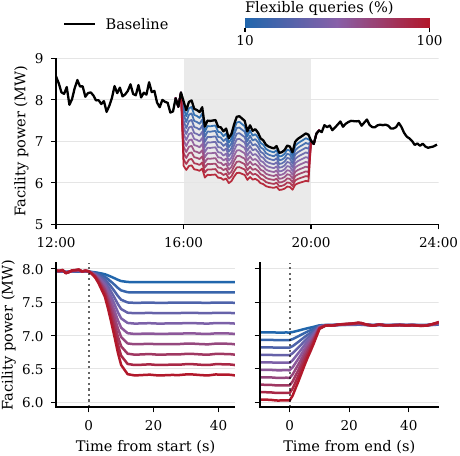}
  \vspace*{-2em}
\caption{Datacenter power control during a four-hour curtailment event.
Top: Sustained power reductions achieved through flexible query routing.
Bottom: Event boundaries showing fast transitions between operating points in the order of seconds.}
  \label{fig:datacenter_curtailment_event}
  \vspace*{-0.5em}
\end{figure}

\paragraph{Practical power curtailment event.}
During periods of high electricity demand or grid disturbances, grid operators may request large loads to curtail their power usage for time periods up to hours. Responding to such curtailment events requires data centers to respond under short notice and sustain the reduction throughout the window, while continuing to serve incoming queries. 

We simulate a power demand curtailment event for a four-hour duration in \Cref{fig:datacenter_curtailment_event} (top half), and inspect the adequacy of our approach for controlling the datacenter facility power. The mean facility power consumption for the baseline allocation throughout this window is $7.28$\,MW. At $\sigma_F\!=\!0.5$, the average curtailed power consumption is $6.7$\,MW and $6.1$\,MW at $\sigma_F\!=\!1$, yielding approximately $2.4$ and $4.8$\,MWh energy reduction during the curtailment event for ensuring grid compliance.

\Cref{fig:datacenter_curtailment_event} (bottom half) also illustrates the power control response transients between distinct operating points in the routing decision space. For the query prompt and output lengths sampled from the Azure trace \cite{stojkovic2025dynamollm}, we observe a transient power response time of approximately $10$s to achieve $90\%$ of the desired power reduction. Thus, query routing across heterogeneous language model instances enables response to grid events within seconds, and can sustain lower power consumption values as query workload and available GPU resources permit. 

\paragraph{GPU availability and resource sharing.}
The available resources for serving inference queries at a datacenter can vary as some GPUs may be allocated to training jobs and other workloads, or may be inaccessible due to maintenance or temporary failures. We therefore examine the power demand flexibility for a given query workload under varying resource availability. 

\begin{figure}[t]
  \centering
  \includegraphics[width=\columnwidth]
    {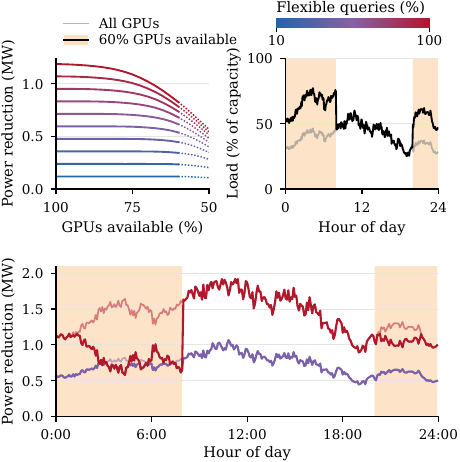}
  \vspace*{-2em}
\caption{Impact of GPU availability and resource sharing on power demand flexibility.
Top: Achievable power curtailment under reduced GPU availability and the resulting utilization during scheduled sharing.
Bottom: Daily power flexibility with GPU sharing, demonstrating resilience to moderate resource constraints.}
  \label{fig:datacenter_sharing_availability}
  \vspace*{-1em}
\end{figure}

\Cref{fig:datacenter_sharing_availability} (top left) demonstrates the average power demand flexibility against the available GPU resources. We observe that moderate resource unavailability preserves most of the demand flexibility, which gradually decays at higher GPU constraints. 
This robustness of the demand flexibility to GPU unavailability reflects the existing headroom available, and reducing the availability reduces the small model pool's capacity to accommodate the flexible query workload, reducing the overall power curtailment capacity.  
Further unavailability eventually leads to violation of capacity and feasibility constraints (\cref{eq:feasible_routing_fraction_range}), leading to queue buildup and service denial for incoming queries. 

We also simulate a scheduled resource sharing scenario, where $40\%$ GPUs are unavailable for inference serving from $8$pm to $8$am.
\Cref{fig:datacenter_sharing_availability} shows the resulting resource utilization and available power flexibility. With half the requests as flexible, we preserve up to $99\%$ of the mean demand flexibility, but we lose a higher fraction of the demand flexibility for the full-flexible query case.

\subsection{Broader Applications and Scope}
\label{subsec:datacenter_broader_applications}

The analytical model for autoregressive inference and power consumption developed in \Cref{sec:inference_power_model} supports designing serving systems at scale. 
Predictions of inference behavior can help guide appropriate choice of number of model instances of each type, deployment comparisons, and GPU-sharing schedules. Along with measured power consumption profiles, these predictions bridge the gap between inference service requirements and feasible power constraints and grid compliance. 
\Cref{tab:datacenter-applications} summarizes potential applications of our approach.

Extending the framework to training regimes or data storage scenarios can enable joint resource planning under common power objectives. For example, datacenters can orchestrate GPU sharing for training jobs, or alternate power sources and battery support when query routing alone cannot satisfy a curtailment target. 
Electricity costs and carbon-intensity goals also support comparing actions by cost and emissions.

Our broad formulation readily extends to multiple models, serving quantization levels, and reasoning budgets, where each choice can be characterized by the associated model instance profiles and query workloads and feasibility. 
For distributed datacenters, local grid constraints may lead to 
routing across regions while respecting network latencies and data-privacy
constraints.

\begin{table}[h]
\vspace*{-2mm}
  \caption{Broader applications of our presented framework.}
  \vspace*{1mm}
  \centering\small
  \begin{tabular}{@{}p{0.31\linewidth}p{0.65\linewidth}@{}}
    \toprule
    Domain & Application\\
    \midrule
    Serving design
      & Model instance pool sizing, deployment comparisons, demand growth headroom.\\
    Power Control
      & Feasible targets, sustained power curtailments,
        and transition planning.\\
    Energy coordination
      & Joint inference/training budgets, storage support,
        and cost or carbon objectives.\\
    Broader routing
      & Allocation across execution choices or locations,
        subject to eligibility constraints.\\
    \bottomrule
  \end{tabular}
  
  \label{tab:datacenter-applications}
  \vspace*{-1em}
\end{table}

\section{Conclusion}
\label{sec:conclusion}

In this work, we investigate the potential of query routing for datacenter power flexibility to support safe grid operation. 
We present a comprehensive analytical model for autoregressive inference and power consumption, allowing us to connect query workload to GPU power demand. Utilizing this, we examine the arising power demand flexibility in datacenter scale operations under practical scenarios. 
More broadly, this work represents an initial step toward the integration of inference systems and power-aware datacenter operations. We envision that insights and framework developed in this work will motivate a large body of research on grid-responsive AI infrastructure, enabling future advances in workload orchestration, power demand management, and reliable integration of large-scale computing into the grid. 

\section*{Acknowledgments}
This work was supported by a PITA Seed Grant, and Bosch RTC Pittsburgh. GJ is supported by an AI2C Seed grant and the US National Science Foundation under grants CCF 2045694, CNS-2112471, CPS-2111751, and ONR N00014-23-1-2149. GQ is supported by NSF Grants 2339112, 2512805, AFOSR Grant FA9550261B281, Jane Street, and CMU Manufacturing Futures Institute.

\bibliography{references}
\bibliographystyle{mlsys2027}

\appendix
\clearpage
\crefalias{section}{appendix}
\crefalias{subsection}{appendix}
\crefalias{subsubsection}{appendix}
\section{Related Work}
\label{app:related_works}

\paragraph{Datacenter demand response and workload flexibility.}
Studies on large flexible loads investigate how power curtailment can accommodate new demand using the existing grid capacity \cite{norris2025rethinking}. Carbon aware scheduling and computing systems utilize temporal flexibility in batched machine learning workloads \cite{carbon_aware_computing_datacenters,carbon_aware_scheduling_clusters_305953}, and geographic workload transfer provides flexibility across distributed locations \cite{limitations_carbonaware_temporal_spatioal_scheduling_10.1145/3627703.3650079}. Zeus and Perseus manage the model training energy through GPU power limits and the clock frequency control \cite{zeus_gpu_energy_285082,perseus_energy_training_10.1145/3694715.3695970}. Carbon Responder coordinates performance aware power control across heterogeneous workloads \cite{xing2023carbonrespondercoordinatingdemand}. Field demonstrations further show that AI workload orchestration can provide sustained curtailment \cite{colangelo_phoenix_2025,williams_grid_responsive_2026}. We study power demand flexibility for online (real-time) inference workloads under varying query length distributions and heterogeneous model deployments.

\paragraph{Power-aware inference.}
Prior works on inference power control utilize hardware settings, serving configurations, and workload placement for achieving power objectives. POLCA \cite{patel2023polcapoweroversubscriptionllm} characterizes LLM power consumption and exploits remaining headroom for datacenter level oversubscription. DynamoLLM \cite{stojkovic2025dynamollm} adapts model instance counts, model parallelism and GPU clock frequencies, while $\mu$-Serve jointly optimizes model multiplexing and frequency scaling \cite{muserve_298496}. PALS \cite{pals_hankendi2026palspowerawarellmserving} combines power-performance models with feedback to coordinate power limits and batching policies, extending to dynamic power targets. TAPAS \cite{tapas_10.1145/3676641.3716025} uses model placements, request routing and serving reconfiguration to manage power and thermal constraints. Our examination of model routing complements these directions and can be composed along with their approaches and objectives.

\paragraph{Measurements of inference energy.}
Empirical studies characterize how inference energy consumption varies across language models, hardware choices, query workloads and deployment configurations. \citet{samsi2023wordswattsbenchmarkingenergy} benchmark LLM inference performance and energy across different GPU configurations. AI Energy Score provides standardized task-level energy comparisons \cite{huggingface2025aienergyscore}, and ML.ENERGY Benchmark \cite{chung2026themlenergybenchmark} measures inference energy consumption under practical scenarios and supports optimization of serving configurations. \citet{luccioni2022estimatingcarbonfootprintbloom} assess the carbon footprint for a 176\,B BLOOM model's life-cycle and measure its deployment energy through a live inference endpoint. These studies benchmark energy consumption at the request level, while we utilize model-instance power consumption estimates to predict how query routing across heterogeneous models can provide power demand flexibility for safe and reliable grid operation.

\paragraph{Analytical models and energy estimates.}
Analytical measurements estimate AI energy consumption across different scales. Scenario-based assessments project electricity demand from assumptions about hardware resources and utilization \cite{devries2023growing}, whereas other approaches estimate per-query energy consumption from model computation and hardware performance \cite{epoch2025howmuchenergydoeschatgptuse}. \citet{oviedo2026energy} connect inference energy to the token throughput, node level power, deployment overheads, query lengths and their estimated datacenter level electricity demand. The 1/W law \cite{onewlaw_chen20261wlawanalyticalstudy} examines how context length, KV-cache concurrency, routing topology and the GPU generation affect inference energy efficiency. Alternatively, our work connects the query arrival rate to inference state and power consumption, which we employ for quantifying power demand flexibility for grid compliance.

\paragraph{Serving systems and performance prediction.}
The performance of inference systems relies on the underlying serving frameworks and their scheduling and memory management policies. PagedAttention in vLLM \cite{vllm_10.1145/3600006.3613165} improves KV-cache memory management through virtual memory and paging, and Sarathi-Serve \cite{sarathi_serve_298679} uses chunked prefill computations for reducing prefill and decode interference and improving resource utilization. Splitwise \cite{splitwise_10.1109/ISCA59077.2024.00019} disaggregates prefill and decode computation across heterogeneous compute resources to better utilize compute and memory-capable GPUs. Vidur \cite{vidur_MLSYS2024_b74a8de4} combines kernel operator profiling and simulation to predict serving performance and evaluate deployment configurations. Our analysis utilizes a one-time configuration specific calibration step which jointly accounts for scheduling policies, deployment configurations and serving framework policies.

\paragraph{Model routing and power control.}
FrugalGPT, Hybrid LLM, and RouteLLM study model selection and
cascades that balance inference cost and response
quality \cite{chen2024frugalgpt,ding2024hybrid,route_llm_ICLR2025_5503a7c6}.
Taking query--model assignments as a given design input, we study how query routing across model instances affects the total power consumption for a given query arrival rate and query and response length distribution. We choose desired allocations to achieve power objectives based on the feasible routing decisions, while respecting capacity constraints and available headroom at model instances. Our formulation connects quality-aware model routing to datacenter power management and control. 

\counterwithin{figure}{section}
\counterwithin{table}{section}
\section{Additional Experimental Results}
\label{app:additional_results}

\subsection{Experimental Setup}
\label{app:experimental-setup}

\paragraph{Hardware and workloads.}
We extend our evaluation across hardware setups to NVIDIA RTX~5090 32\,GB, Quadro RTX~8000 48\,GB, and Tesla V100 32\,GB GPUs. Models are served using bfloat16 on H100 and RTX-5090, and RTX~8000 and V100 utilize float16. Each GPU hosts one vLLM serving instance with chunked prefills, which is further generalized to tensor parallel configuration for Llama 70B on four H100 GPUs.  The models used in our evaluations for single GPU serving configurations span 1B--24B parameters.

The query and response length distribution for our analysis is sampled from fixed length control settings (128/128, 256/256, 512/512, 1024/1024, 2048/2048, and 6144/1024 tokens) and varying length practical distributions (ShareGPT, HumanEval, arXiv summarization). The GPU board power consumption is measured via the Nvidia Management Library.

\begin{table}[h]
\caption{Key quantities and their notation as presented in the LLM Inference and Power Consumption model, \Cref{sec:inference_power_model}.}
\centering\small
\begin{tabular}{@{}p{0.25\linewidth}p{0.69\linewidth}@{}}
\toprule
Notation & Description\\
\midrule
$\lambda$
& Query arrival rate (requests/s)\\
$b$ & request concurrency\\
$b_{\max}$ & maximum request concurrency\\
$\bar p,\bar d$ & Mean prompt and output lengths (tokens)\\
$\text{chunk}_{\mathrm{pre}}$ & number of prefill tokens in one chunk\\
$s$ & approximate token iterations per request\\
$\tau(b)$ & Mean time between output tokens (s)\\
$\widehat T(b)$ & Estimated mean request residence time (s)\\
$\widehat\mu(b),\widehat C(b_{\max})$ & Estimated request throughput and serving capacity (requests/s)\\
$P(b),P_{\mathrm{idle}}$ & Model Instance power profile, idle power (W)\\
\bottomrule
\end{tabular}
\label{tab:power-notation}
\end{table}

\begin{figure*}[h]
  \centering
  \includegraphics[width=0.9\textwidth]
    {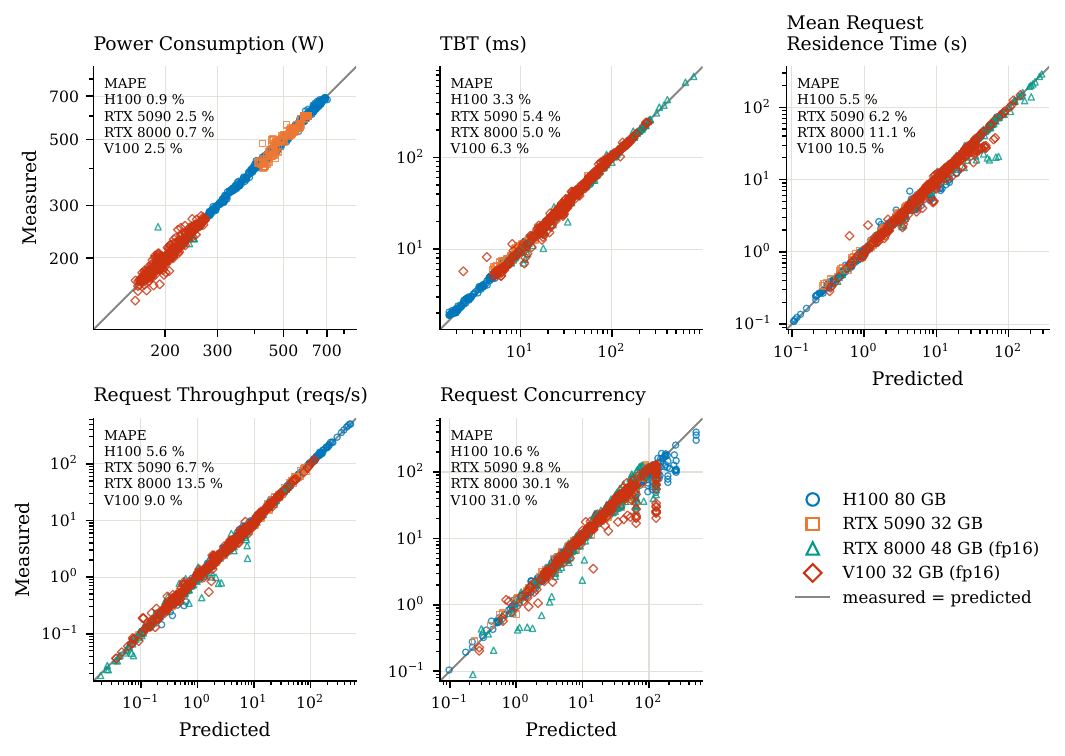}
  \caption{Single model instance predictions across hardware choices. Across query length distributions, language models and serving framework configurations, our predictions closely track the actual measured values, demonstrating accuracy throughout the examined scenarios.}
  \label{fig:app_single_gpu_exps}
\end{figure*}

\subsection{Evaluation and Discussion}
\label{app:cross-gpu-validation}

\Cref{fig:app_single_gpu_exps} compares the predictions and the measured values across the four hardware platforms. \Cref{tab:app_single_gpu_accuracy} reports the mean absolute percentage errors (MAPE) for the modeled quantities.

\paragraph{Prediction sensitivity near max throughput.}
Mean request concurrency estimation shows errors exceeding $10\%$ MAPE in some cases. The large errors are majorly observed in long sequence configurations and concurrency when GPUs are near capacity. The fixed point estimate in \cref{eq:occupancy_estimator} becomes sensitive as the $\lambda s t_1\!\to\!1$, resulting in magnification of small prediction errors in time-between-tokens estimates. However, the power consumption remains relatively unaffected by prediction errors in number of resident requests as power consumption profiles plateau quickly with increasing concurrency (\Cref{fig:power_consumed_vs_occupancy}). 

\paragraph{Tensor-parallel deployment.}
We additionally evaluate the Llama-3.1-70B language model served with tensor parallel configuration across four H100 GPUs, and model the inference and power consumption behavior for this model instance. Our evaluation covers nine query length distributions, and we observe a $2.8\%$ error in power consumption estimates and $4.3\%$ error in maximum throughput estimates across the query length distributions. Furthermore, the TBT measurements are accurate to $2.8\%$ of the measured values. These results demonstrate that our analytical model for inference and power consumption generalizes well to large models served across multiple GPUs as a single model instance. 

\paragraph{Memory-limited concurrency.}
KV cache capacity imposes a limit in addition to TBT and latency constraints that determine the operating concurrency cap $b_{\max}$.  We estimate this memory-limited concurrency as
\begin{equation*}
b_{\max}\leq
\left\lfloor\frac{K}{\bar\ell}\right\rfloor,
\qquad
\bar\ell\approx
\frac{\mathbb E[pd+d(d+1)/2]}{\mathbb E[d]},
\end{equation*}
where $K$ denotes the KV-cache capacity in tokens, and $p$ and $d$ denote individual prompt and output lengths of queries. The estimate $\bar\ell$ represents the average context length weighted by the number of decode tokens, accounting for longer generations and context growth.

\paragraph{Time to Idle.}
When new requests are no longer admitted, a model instance continues processing resident requests until all sequence generations are complete. This transition typically takes order of seconds and depends primarily on the remaining output lengths and the number of active requests. Characterizing this interval helps capture the transient behavior of inference systems as they transition from active serving to idle operation.

\paragraph{Power Estimation.}
For each hardware configuration, language model and query length distributions, we model the power consumption as the sum of idle power and concurrency-dependent dynamic component (\cref{eq:model_instance_power}). We measure the idle model instance power directly from the model instance with no requests being served. To characterize the dynamic power, we measure average power across varying values of request concurrency ($b$), and calibrate candidate curves to the measured values and select the one with lowest prediction error on held out values. The resulting curves exhibit sharp increase in power consumption as number of active sequences grows, followed by a gradual plateau as GPU utilization approaches saturation. For H100 GPU evaluations, the measured curves achieve an average prediction error of approximately 1\,\%.

\subsection{Additional Routing Results}
\label{app:fleet-routing}

\begin{figure*}[p]
  \centering

  \begin{minipage}[t]{0.45\textwidth}
    \centering
    \includegraphics[
      width=\linewidth,
      height=0.84\textheight,
      keepaspectratio
    ]{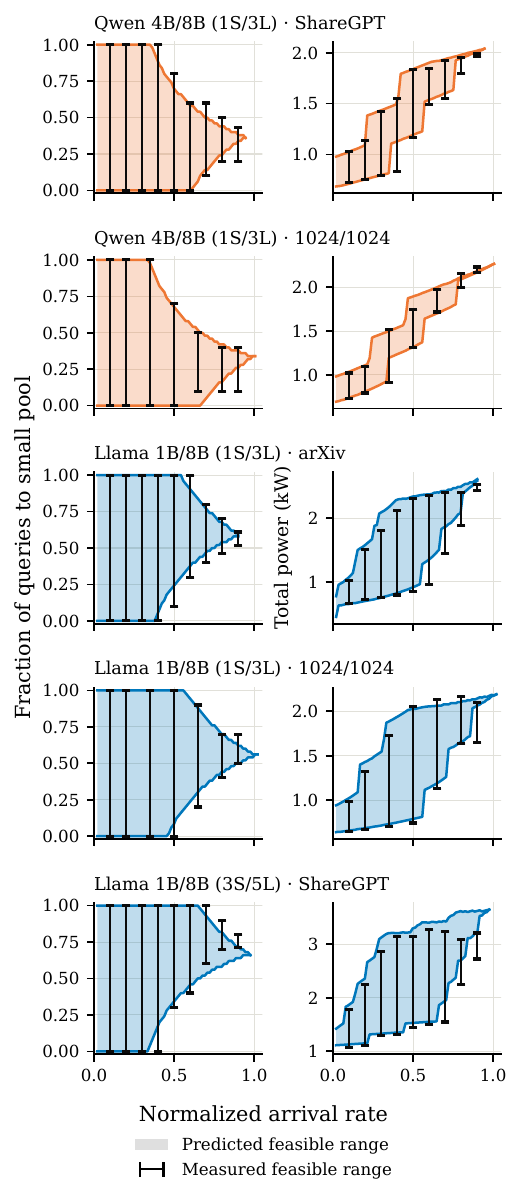}

    \caption{Capacity-feasible routing decisions (left, \cref{eq:feasible_routing_fraction_range}) and corresponding total-power consumption (right, \cref{eq:routing_total_power}) across query arrival rates and model--workload configurations. Shaded regions show analytical predictions, bars depict measured feasible ranges. S/L denotes the number of
    small/large model instances. The routing interval contracts with
    increasing arrival rates across scenarios.}
    \label{fig:app_routing_window}
  \end{minipage}
  \hfill
  \begin{minipage}[t]{0.53\textwidth}
    \centering
    \includegraphics[
      width=\linewidth,
      height=0.84\textheight,
      keepaspectratio
    ]{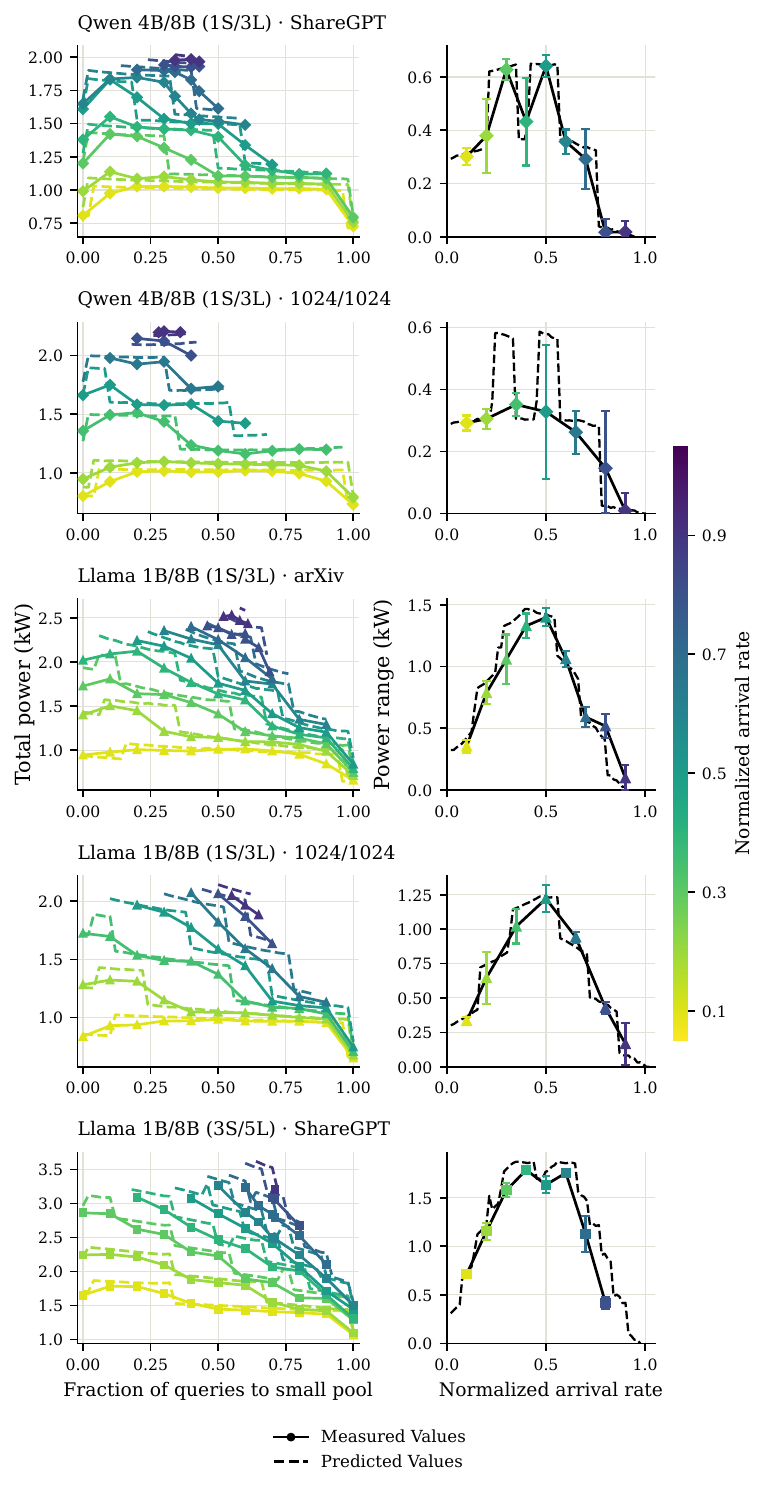}

    \caption{Total power consumption via query routing for the
    configurations in \Cref{fig:app_routing_window}. Left: measured and predicted total power (\cref{eq:routing_total_power}) versus the fraction of flexible requests ($f$) assigned to the small model instance, with varying query arrival rate. Right: measured and predicted power consumption range (\cref{eq:feasible_routing_power}) over feasible routing allocations.}
    \label{fig:app_routing_power}
  \end{minipage}

\end{figure*}

\Cref{fig:app_routing_window,fig:app_routing_power} extend the investigation of power demand flexibility through query routing (\Cref{fig:feasible_routing_power,fig:power_range_vs_load}) to other model and query length distribution choices. We also include analysis of an eight model instance pool, with three small and five large models (3S/5L). As observed in \Cref{sec:routing_power_control}, the feasible routing decision range contracts as the query arrival rate increases, and power flexibility is generally largest at intermediate arrival rates. 
Across all the configurations and measurements, the total power consumption predictions achieve $3.3\%$ error, consistent with the low prediction errors for single model instance estimates.

\subsection{Feedback Power Control}
\label{app:feedback_control}

Our analytical model selects routing decisions for power control based on the average query arrival rate and query length distribution. To further augment decision making, we implement a PID (Proportional Integral Derivative) feedback controller which adjusts routing decisions based on real-time power consumption measurements.

\paragraph{Controller implementation.}
The PID controller updates the routing fraction ($f$) decision using NVML measurements at intervals of $1$sec. The normalized control error used for feedback is 
\begin{equation*}
  e_k=\frac{\widetilde P_k-P_{\mathrm{target},k}}{B_{\mathrm{ref}}},
\end{equation*}
where $\widetilde P_k$ is the measured total power and $B_{\mathrm{ref}}$ is the measured power range. A positive error implies a higher than desired power consumption, prompting an increase in the fraction of queries routed to the small model instance pool. We use calibrated gains $k_p=0.154$, $k_i=0.143$, and $k_d=0.077$ for the standard PID controller. We further apply a $40$W deadband region to limit correction for small errors and fluctuations in power measurement.

\paragraph{Evaluation protocol.}
We evaluate both the analytical model and the PID controller on Llama-1B/8B (1S/3L) setup under varying query length distributions. We evaluate multiple target power reductions, and measure the stable behavior. We quantify the performance through mean-power error, which is the absolute difference between the target and the average measured power under steady operation. We also measure the standard deviation at steady state and the response time ($t_{90}$) required to achieve $90\%$ of the desired power reduction.

\paragraph{Results.}
Table~\ref{tab:app_feedback_control_comparison} describes that the PID feedback improves the mean power value achieved as compared to the analytical model ($14$ vs $34$\,W error) across the examined configurations. However, PID controller leads to higher power fluctuations due to continuous real-time correction and susceptibility to power measurement noise as compared to the analytical model which prescribes a fixed choice of operating routing decisions. Furthermore, PID controller also demonstrates a higher median $t_{90}$ response time of $12$\,s as compared to $6.5$\,s for the analytical model.

\paragraph{Implications.}
Both the fixed routing decision through analytical model and real-time feedback control are complementary approaches. The fixed decision for power curtailment enables a rapid initial allocation, which can be composed with a slower feedback loop which subsequently corrects residual errors and also adapts to workload variations and evolving conditions under real deployment scenarios.

\begin{table*}[t]
  \centering\small
    \caption{Single model instance prediction errors across all evaluated
  models and workloads on hardware setups. Mean Absolute Percentage Error (MAPE) presented for each case.}
  \vspace*{1mm}
  \setlength{\tabcolsep}{9pt}
  \begin{tabular}{@{}lrrrr@{}}
    \toprule
    Quantity & H100 & RTX~5090 & RTX~8000 & V100 \\
    \midrule
    Power consumption at concurrency $b$, $P(b)$
      & 0.9 & 2.5 & 0.7 & 2.5 \\
    Power consumption at query arrival rate $\lambda$, $\widehat P(\lambda)$
      & 5.3 & 6.9 & 13.3 & 8.8 \\
    Token iteration time, $\tau(b)$
      & 3.3 & 5.4 & 5.0 & 6.3 \\
    Request residence time, $\widehat T(b)$
      & 5.5 & 6.2 & 11.1 & 10.5 \\
    Request Throughput, $\widehat\mu(b)$
      & 5.6 & 6.7 & 13.5 & 9.0 \\
    Request Concurrency, $\widehat b(\lambda)$
      & 10.6 & 9.8 & 30.1 & 31.0 \\
    Serving capacity
      & 6.0 & 7.2 & 10.0 & 9.9 \\
    KV-cache concurrency ceiling
      & 2.8 & 1.7 & 7.7 & 4.6 \\
    \bottomrule
  \end{tabular}
  \label{tab:app_single_gpu_accuracy}
\end{table*}

\begin{table*}[t]
\centering\small
  \caption{Comparison of open-loop analytical model based power control and PID feedback control across analyzed scenarios. Mean-power error is the absolute difference between average measured power and the target; power standard deviation is the standard deviation of measured power values; response time $t_{90}$ denotes time to achieve 90\% of the desired power reduction. PID controller demonstrates lower mean-power error, but at the cost of higher fluctuations and response times.}
\vspace{1mm}
  \setlength{\tabcolsep}{7pt}
  \begin{tabular}{@{}lrrrrrr@{}}
    \toprule
    & \multicolumn{2}{c}{Mean-power error (W)} & \multicolumn{2}{c}{Power std.\ dev.\ (W)} & \multicolumn{2}{c}{Response time $t_{90}$ (s)} \\
    \cmidrule(lr){2-3}\cmidrule(lr){4-5}\cmidrule(lr){6-7}
    Controller & Median & 90th pct. & Median & 90th pct. & Median & 90th pct. \\
    \midrule
    Analytical model, open loop & 34 & 131 &  27 & 136 &  6.5 & 16.5 \\
    PID feedback                & 14 &  85 & 108 & 194 & 12.0 & 18.6 \\
    \bottomrule
  \end{tabular}

  \label{tab:app_feedback_control_comparison}
\end{table*}

\subsection{Grid-Service Timescales and Datacenter Results}
\label{app:grid-timescales}

\begin{figure*}[h]
  \centering
  \includegraphics[width=0.98\textwidth]
    {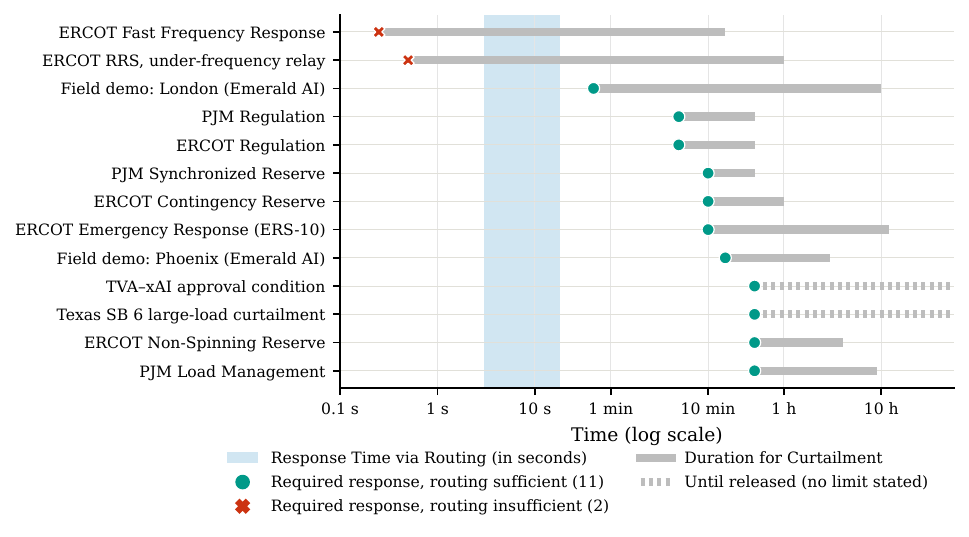}
  \caption{Response timescales under query routing for power control, with respect to grid-service and curtailment requirements. The initial response time (in seconds) for query routing is fast enough to accommodate majority of the demand flexibility requirements, which can be further coupled with approaches like battery/UPS systems for fast response and sustained power control.}
  \label{fig:app-grid-timescales}
\end{figure*}

\paragraph{Grid programs and response requirements.}
We present popular grid operation requirements and power curtailment policies in \Cref{tab:app-grid-programs}, and summarize the timescales of operation in \Cref{fig:app-grid-timescales}. We consider formal grid service agreements, site-specific curtailment conditions and field demonstrations for this discussion. 

\paragraph{Query Routing Response Timeframes.}
Across our evaluations on varying query workloads, the measured $t_{90}$
span is approximately $3\!-\!18$\,s. These response times are considerably faster than most minute-hour scale deadlines described in \Cref{tab:app-grid-programs}, whereas some fast-frequency response agreements require sub-second response latencies. Query routing alone cannot provide response in sub-second latencies, but coupled with other short-term quick response solutions like battery/UPS systems, our approach can sustain all of the presented grid operation policies across timescales.

\paragraph{Absorbing surplus power.}
Our main analysis and experiments focus on power curtailment applications, but our framework can also guide demand surge to absorb surplus renewable power generation in the grid \cite{eia2025curtailment,nationalgrid2018dtu}. The serving and power consumption models can estimate the available headroom for additional power consumption by utilizing larger model instances subject to capacity and feasibility constraints. 
Prior work has demonstrated secondary GPU workloads for power smoothing \cite{choukse2025power}, along with deferred batch workload scheduling to attain higher power consumption values.

\begin{table*}[p]
  \centering\footnotesize
    \caption{Grid-service requirements and contextual examples
  accompanying \Cref{fig:app-grid-timescales}. 
  Holding capability, market commitments, cumulative deployment
  obligations, and demonstrated event durations are identified
  separately. Response speed alone does not establish service
  qualification or the ability to deliver the required magnitude.}
  \vspace*{2mm}
  \setlength{\tabcolsep}{4pt}
  \renewcommand{\arraystretch}{1.12}
  \begin{tabular}{@{}>{\raggedright\arraybackslash}p{0.28\textwidth}
                    >{\raggedright\arraybackslash}p{0.29\textwidth}
                    >{\raggedright\arraybackslash}p{0.37\textwidth}@{}}
    \toprule
    Program or Guideline & Response Requirement
      & Duration and Objectives \\
    \midrule
    \multicolumn{3}{@{}l}{\textit{Grid services and demand response}} \\
    \addlinespace[2pt]
    ERCOT Fast Frequency Response
      \citep{ercot_operating_guide_2026}
      & Full response within 15 cycles (0.25\,s at 60\,Hz).
      & Sustain for 15\,min or until recall, whichever occurs first. \\
    \addlinespace[3pt]
    ERCOT RRS, high-set under-frequency relay
      \citep{ercot_operating_guide_2026,ercot_protocols_2026}
      & Interrupt load within 30 cycles (0.5\,s).
      & At least 1\,h capability for this resource category following restoration instructions. \\
    \addlinespace[3pt]
    ERCOT Regulation \citep{ercot_protocols_2026}
      & Respond to control signals within 5\,s; deliver capacity within 5\,min.
      & At least 30\,min duration capability with continuous signal tracking. \\
    \addlinespace[3pt]
    PJM Regulation \citep{pjm_manual12_2026,pjm_manual11_2026}
      & Respond immediately and reach bid capability within 5\,min.
      & Half-hour commitments with continuous tracking and
        performance scoring. \\
    \addlinespace[3pt]
    ERCOT Emergency Response, ERS-10
      \citep{ercot_protocols_2026}
      & Meet deployment obligations within 10\,min of dispatch.
      & Sustain until recall. Cumulative obligations are
        24\,h in December--March contract periods and
        12\,h in other terms. \\
    \addlinespace[3pt]
    ERCOT Contingency Reserve \citep{ercot_operating_guide_2026}
      & Deploy within 10\,min.
      & At least 1\,h sustained capability under the cited
        operating guide. \\
    \addlinespace[3pt]
    PJM Synchronized Reserve \citep{pjm_manual11_2026}
      & Delivering assigned response within 10\,min.
      & Maintain the achieved response until the event ends
        or 30\,min from event start, whichever is earlier. \\
    \addlinespace[3pt]
    ERCOT Non-Spinning Reserve \citep{ercot_protocols_2026}
      & Deploying reserves within 30\,min.
      & At least 4\,h qualification capability under described
        rules, load deployment is maintained until recall. \\
    \addlinespace[3pt]
    PJM Load Management \citep{pjm_manual18_2026,pjm_manual13_2026}
      & Full reduction within 30\,min by default,
        exceptions allow 60 or 120\,min.
      & Dispatch-dependent duration until release. \\
    \midrule
    \multicolumn{3}{@{}l}{\textit{Site-specific curtailment conditions}} \\
    \addlinespace[2pt]
    PUCT Docket 58881 \citep{puct_58881_2026}
      & Full curtailment within 30\,min of instruction,
        10\,min if capability is confirmed beforehand.
      & ERCOT provides 60\,min notice before issuing
curtailment instruction when applicable.
Re-energization requires authorization. \\
    \addlinespace[3pt]
    TVA--xAI, reported approval condition \citep{saad_tva_xai_2024}
      & Ability to curtail on a 30-minute TVA request.
      & Site-specific condition. \\
    \midrule
    \multicolumn{3}{@{}l}
      {\textit{Field demonstrations}} \\
    \addlinespace[2pt]
    Phoenix, Emerald AI \citep{colangelo_phoenix_2025}
      & Utility-specified 15-minute ramp.
      & Demonstrated 25\% reduction held for 3\,h. \\
    \addlinespace[3pt]
    London, Emerald AI \citep{williams_grid_responsive_2026}
      & Demonstrated approximately 30\% reduction within 40\,s.
      & Separate sustained-curtailment tests held
        10\%--40\% reductions for 2--10\,h. \\
    \bottomrule
  \end{tabular}
  \label{tab:app-grid-programs}
\end{table*}

\end{document}